\documentclass[conference]{IEEEtran}

\makeatletter

\def\ps@IEEEtitlepagestyle{%
  \def\@oddfoot{\mycopyrightnotice}%
  \def\@evenfoot{}%
}
  \gdef\mycopyrightnotice{}
\makeatother

\usepackage{blindtext}
\usepackage{eso-pic}
\IEEEoverridecommandlockouts
\usepackage{tikz}
\usepackage{pgfplots}
\usepackage{booktabs}
\usepackage{array}
\pgfplotsset{compat=1.18}

\pgfplotsset{compat=1.18}
\usepackage{cite}
\usepackage{amsmath,amssymb,amsfonts}
\usepackage{algorithmic}
\usepackage{graphicx}
\usepackage{textcomp}
\usepackage{xcolor}
\def\BibTeX{{\rm B\kern-.05em{\sc i\kern-.025em b}\kern-.08em
    T\kern-.1667em\lower.7ex\hbox{E}\kern-.125emX}}

\begin{document}
\title{\vspace*{1cm} Prompt-Based Clarity Evaluation and Topic Detection in Political Question Answering\\
\thanks{}
}

\author{\IEEEauthorblockN{ Lavanya Prahallad}
\IEEEauthorblockA{\textit{Research Spark Hub Inc.}\\
%\textit{International Institute of Information Technology}\\
Dublin, CA, USA \\
lavanya@researchsparkhub.com}
\\
\IEEEauthorblockN{Sai Utkarsh Choudarypally}
\IEEEauthorblockA{\textit{Emerald High School} \\
Dublin, CA, USA \\
choudarypally.sai2009@gmail.com}
\and
\IEEEauthorblockN{Pragna Prahallad}
\IEEEauthorblockA{\textit{Emerald High School} \\
Dublin, CA, USA \\
pragnaprahallad@gmail.com}
\\
\IEEEauthorblockN{Pranathi Prahallad}
\IEEEauthorblockA{\textit{Emerald High School} \\
Dublin, CA, USA \\
pranathiprahallad@gmail.com}
}

\maketitle
%\conf{\textit{  VI. International Conference on Electrical, Computer and Energy Technologies (ICECET 2026) \\ 
%6-8 July 2026, Rome-Italy}}
\begin{abstract}
Automatic evaluation of large language model (LLM) responses requires not only factual correctness but also clarity, particularly in political question–answering. While recent datasets provide human annotations for clarity and evasion, the impact of prompt design on automatic clarity evaluation remains underexplored.
In this paper, we study prompt-based clarity evaluation using the CLARITY dataset from the SemEval-2026 shared task. We compare a GPT-3.5 baseline provided with the dataset against GPT-5.2 evaluated under three prompting strategies: simple prompting, chain-of-thought prompting, and chain-of-thought with few-shot examples. Model predictions are evaluated against human annotations using accuracy and class wise metrics for clarity and evasion, along with hierarchical exact match.
Results show that GPT-5.2 consistently outperforms the GPT-3.5 baseline on clarity prediction, with accuracy improving from 56\% to 63\% under chain-of-thought with few-shot prompting. Chain-of-thought prompting yields the highest evasion accuracy (34\%), though improvements are less stable across fine grained evasion categories. We further evaluate topic identification and find that reasoning based prompting improves accuracy from 60\% to 74\% relative to human annotations.
Overall, our findings indicate that prompt design reliably improves high level clarity evaluation, while fine grained evasion and topic detection remain challenging despite structured reasoning prompts.
\end{abstract}

%\copyrightnotice{XXX-X-XXXX-XXXX-X/XX/\$XX.00 ©20XX IEEE}

\begin{IEEEkeywords}
Clarity evaluation, Prompt engineering, Political Question–Answering, Large language models, Chain-of-thought prompting
\end{IEEEkeywords}

\section{Introduction}
Large language models (LLMs) are increasingly deployed in high stakes domains such as politics, healthcare, and education. In these settings, model outputs must be not only factually correct but also clear, interpretable, and responsive to the intent of the input. Automatically evaluating response clarity, however, remains challenging due to its subjective and context-dependent nature, particularly in open-ended question answering scenarios \cite{celikyilmaz2020evaluation, gehrmann2021gem}.

Recent work has begun to address this challenge through datasets with human annotated clarity judgments. One such resource is the CLARITY dataset released as part of the SemEval 2026 shared task \cite{clarity-semeval}. CLARITY provides hierarchical annotations that capture both high level clarity judgments and fine grained evasion behaviors in political question answering. These annotations enable direct comparison between human judgments and model predictions \cite{clarity-semeval, thomas2024clarity}.

A common assumption in LLM research is that reasoning oriented prompt design, such as chain-of-thought prompting, improves model performance across tasks. While prior work demonstrates the effectiveness of reasoning based prompting for structured inference and multi step reasoning \cite{wei2022cot, kojima2022zero, zhou2023least}, its impact on subjective evaluation tasks such as clarity and evasion detection remains underexplored. In particular, it is unclear whether increased prompt complexity consistently improves alignment with human judgments across hierarchical labeling schemes.

In this work, we empirically examine the effect of prompt design on automatic clarity evaluation in political question answering. Using the CLARITY dataset, we compare simple prompting, chain-of-thought prompting, and chain-of-thought with few-shot examples applied to GPT-5.2, and benchmark performance against GPT-3.5 predictions provided with the dataset. In addition to clarity and evasion prediction, we evaluate topic identification accuracy, as accurate topic recognition is essential for interpreting whether a response appropriately addresses a question.

Our results show that GPT-5.2 consistently outperforms the GPT-3.5 baseline on clarity prediction, with accuracy improving from 56\% to 63\% under chain-of-thought prompting with few-shot examples. Topic identification accuracy improves from 60\% to 74\% with reasoning based prompting. However, gains for fine grained evasion classification are less stable, indicating that increased prompt complexity does not uniformly improve performance across evaluation dimensions. Overall, this work presents a reproducible, prompt-based evaluation framework and a controlled empirical comparison of prompting strategies, demonstrating that reasoning oriented prompts yield uneven benefits across hierarchical clarity, evasion, and topic detection tasks.

\section{Clarity and Evasion Labeling}

We use the CLARITY dataset released as part of the SemEval-2026 shared task on clarity and ambiguity in political question–answering. The dataset consists of political question–answer pairs annotated by three human annotators with both clarity-level and evasion-level labels, enabling direct comparison between human judgments and automated predictions.

Each question-answer pair in the dataset is annotated independently by three human annotators. These annotations include both clarity-level and evasion-level labels, enabling direct comparison between human judgment and automated predictions.

\subsection{Dataset Structure}

Each data instance consists of a question-answer pair. Questions are drawn from political interviews or discussions, and answers correspond to responses given by speakers.

The dataset contains two related types of annotations:
\begin{itemize}
    \item clarity-level labels, which indicate whether a response answers the question clearly, ambiguously, or not at all;
    \item evasion-level labels, which provide a finer grained description of how a response avoids or partially addresses the question.
\end{itemize}

\subsection{Clarity and Evasion Taxonomy}

The CLARITY dataset follows a hierarchical labeling taxonomy based on the framework proposed by Thomas et al.\ \cite{thomas2024clarity}. At the top level, each response is assigned one of three clarity-level labels: 

\begin{itemize}
    \item Clear Reply: the response directly answers the question;
    \item Ambivalent Reply: the response is related to the question but is indirect, unclear, or incomplete;
    \item Clear Non-Reply: the response clearly does not answer the question.
\end{itemize}

Each clarity-level label is further refined using evasion-level labels that describe how the response behaves with respect to answering the question.

\subsection{Dataset Splits and Pre-processing}

The original dataset is divided into a training set with 3,448 question-answer pairs and a test set with 308 question-answer pairs.

We downloaded the dataset from the official CLARITY shared task repository and processed it locally. The data was converted into CSV format, and duplicate question-answer pairs were removed. After cleaning, the dataset used in our experiments contains 2,061 unique question-answer pairs.

\subsection{Human Annotations and Baseline Evaluation}

The human annotations provided by the three annotators are used as ground truth in our experiments. As a baseline for automatic clarity prediction, we compare these labels against predictions generated by ChatGPT-3.5 \cite{openai2023gpt35}.

Evaluation is performed on 2,047 scored instances. Table~\ref{tab:gpt35_baseline} reports overall accuracy as well as class wise precision, recall, and F1-score for clarity-level prediction.

\begin{table}[t]
\centering
\caption{Baseline clarity prediction performance of ChatGPT-3.5}
\label{tab:gpt35_baseline}
\begin{tabular}{lccc}
\toprule
\textbf{Class} & \textbf{Precision} & \textbf{Recall} & \textbf{F1-score} \\
\midrule
Clear Reply        & 0.57 & 0.55 & 0.56 \\
Ambivalent Reply   & 0.59 & 0.64 & 0.61 \\
Clear Non-Reply    & 0.36 & 0.27 & 0.31 \\
\midrule
\textbf{Overall Accuracy} & \multicolumn{3}{c}{\textbf{0.56}} \\
\bottomrule
\end{tabular}
\end{table}

The results show that ChatGPT-3.5 performs reasonably on Clear Reply and Ambivalent Reply instances but struggles with Clear Non-Reply cases. This baseline highlights the difficulty of automatic clarity evaluation and motivates our investigation into whether improved prompting strategies can lead to better alignment with human judgments.

\section{Topic detection}

Accurate topic detection identifies the primary issue addressed by a political question and provides essential context for clarity and evasion evaluation. Because political questions often reference multiple issues, assigning an incorrect topic can lead to misleading judgments about response clarity or evasiveness.
In this study, topic detection is performed in two stages: human annotators first assign gold-standard topic labels, and ChatGPT then predicts topics using different prompt formulations, with predictions evaluated against human annotations. A fixed set of topic categories and representative lexical cues is used to ensure consistency across annotation and prediction.

\begin{table}[t]
\centering
\caption{Topic definitions and representative detection cues}
\label{tab:topic_definitions}
\renewcommand{\arraystretch}{0.85}
\begin{tabular}{>{\raggedright\arraybackslash}p{2.3cm} >{\raggedright\arraybackslash}p{4.1cm}}
\toprule
\textbf{Topic} & \textbf{Representative Detection Cues} \\
\midrule
Abortion & abortion, pro-choice, pro-life, reproductive rights \\
Climate Change and Energy & climate change, emissions, energy, fossil fuels, renewables \\
COVID-19 & COVID-19, pandemic, masks, vaccines, lockdowns \\
Crime and Policing & crime, police, law enforcement, public safety, reform \\
Economy & taxes, inflation, jobs, minimum wage, cost of living \\
Education & schools, colleges, tuition, student loans, teachers \\
Elections and Voting & elections, voting, voter ID, mail-in ballots \\
Foreign Policy & diplomacy, sanctions, NATO, Ukraine, China, Russia \\
Gun Control & guns, Second Amendment, background checks \\
Healthcare & healthcare, insurance, Medicare, ACA, public health \\
Immigration & border, asylum, migrants, citizenship, deportation \\
LGBTQ+ Rights & LGBTQ+, gender identity, sexual orientation \\
War and Security & war, military, national security, nuclear weapons \\
Other & no dominant topical signal \\
\bottomrule
\end{tabular}
\end{table}

\subsection{Human Topic Annotation}

Human topic annotation was performed to create a gold-standard reference for evaluation. Each question was independently analyzed and assigned a single dominant topic based on the primary issue being addressed.

The annotation process followed these principles:

\begin{itemize}
    
    \item \textbf{Keyword guided review:} Questions were examined for topical cues such as references to elections, military activity, country names, energy resources, or public health. Keywords served as indicators rather than definitive labels.
    
    \item \textbf{Question level isolation:} Each question was annotated as a standalone unit. Context from neighboring questions was intentionally ignored to avoid topic leakage across instances.
    
    \item \textbf{Dominant topic selection:} When multiple topics were present, annotators selected the most prominent issue based on emphasis and relevance. Secondary topics were not encoded.
\end{itemize}

\section{Experimental Methodology}

This section describes the experimental setup used to evaluate automatic clarity, evasion, and topic detection. We detail the evaluation pipeline, prompting strategies, task definitions, and metrics. All quantitative results are reported in Section~V.

\subsection{Evaluation Pipeline}

All experiments were conducted in a reproducible Google Colab environment using a fixed evaluation pipeline. The objective of this pipeline is to isolate the effect of prompt design on model alignment with human annotations while holding all other factors constant.

For each experiment, the cleaned CLARITY dataset is loaded and a single prompt configuration is applied uniformly to all question–answer pairs. All prompt-based experiments use GPT-5.2. For each instance, the model produces three outputs: a clarity label, an evasion label, and a topic label. These predictions are evaluated by direct comparison with human annotated ground truth.

\subsection{Prompting Strategies for Clarity and Evasion Labels}

We evaluate two primary prompting strategies with GPT-5.2: simple prompting and chain-of-thought (CoT) prompting. Both strategies use the same dataset, label taxonomy, and evaluation metrics. They differ only in whether the prompt explicitly guides the model through a structured reasoning process before assigning labels.

\subsubsection{Simple Prompt}

The simple prompt instructs the model to assign clarity and evasion labels directly, without requiring explicit intermediate reasoning steps. This configuration serves as a minimal prompting baseline for GPT-5.2.

The exact prompt used is shown below:

\begin{quote}
\textit{You are a data quality evaluator. Analyze the interaction below.}

\medskip
\textbf{Interview Question:} \\
``\{\{interview\_question\}\}''

\medskip
\textbf{Interview Answer:} \\
``\{\{interview\_answer\}\}''

\medskip
Each response must be labeled at two levels:
\begin{itemize}
    \item Level 1 (Clarity): Clear reply, Ambivalent, or Clear non-reply
    \item Level 2 (Evasion): a valid evasion label conditioned on the clarity label
\end{itemize}

Choose one clarity label and one valid evasion label.
\end{quote}

\subsubsection{Chain-of-Thought Prompt}

The chain-of-thought prompt guides the model through a structured decision process. The model is instructed to identify the core request of the question, assess whether the answer satisfies that request, and then assign clarity and evasion labels consistent with this reasoning. 
The exact prompt used is shown below:

\begin{quote}
\textit{You are an expert annotator for a two level clarity and evasion taxonomy.}

\medskip
\textbf{Interview Question:} \\
``\{\{interview\_question\}\}''

\medskip
\textbf{Interview Answer:} \\
``\{\{interview\_answer\}\}''

\medskip
\textbf{Decision Procedure:}
\begin{enumerate}
    \item Identify the core request of the question.
    \item Determine whether the answer provides the requested information (Yes, Partial, or No).
    \item Assign a clarity label.
    \item Assign a valid evasion label consistent with the clarity label.
\end{enumerate}
\end{quote}

\subsection{Prompts for Topic Prediction}

The exact prompt formulations used for topic prediction are shown below:

\begin{quote}
\textbf{Simple:} \textit{You are a political discourse analyst. Identify the primary topic discussed in the interaction below. Assign the most appropriate topic label based on the content of the question and answer.}
\end{quote}

\begin{quote}
\textbf{Reasoning:} \textit{You are a political discourse analyst. First determine what the question is primarily asking about, then analyze how the answer addresses it. Think step by step and assign the most appropriate topic label for the interaction.}
\end{quote}

\subsection{Tasks Evaluated}

We evaluate model performance on three related but distinct tasks:

\paragraph{Clarity Labeling}
Clarity labeling assesses whether a response fully answers the question, partially addresses it, or does not answer it at all. Labels follow the three-way clarity taxonomy defined in the CLARITY dataset.

\paragraph{Evasion Labeling}
Evasion labeling provides a fine grained characterization of how a response avoids answering the question. Evasion labels are conditioned on the clarity label and capture strategies such as deflection, generalization, or refusal.

\paragraph{Topic Identification}
Topic identification assigns a single dominant political topic to each question–answer pair from a fixed set of categories. Topic labels provide contextual grounding for clarity and evasion judgments by indicating what issue the response is expected to address.

\subsection{Evaluation Metrics}

Model performance is evaluated using standard classification metrics. Accuracy measures the proportion of instances for which the predicted label matches the human annotation. Because label distributions are imbalanced, we also report precision, recall, and F1-score for each class.

For clarity and evasion jointly, we report hierarchical exact match accuracy, which measures whether both labels are predicted correctly for the same instance. This metric provides a stricter measure of consistency across the two level labeling structure.

\section{Results}

This section reports experimental results for clarity labeling, evasion labeling, and topic identification by comparing model generated predictions against human annotations. Results are organized to highlight how different prompting strategies affect performance across these related evaluation tasks.

\begin{table}[h]
\centering
\caption{Overall accuracy for clarity and evasion prediction}
\label{tab:overall_accuracy}
\begin{tabular}{|l|c|c|}
\hline
Model / Prompt Strategy & Clarity Accuracy & Evasion Accuracy \\
\hline
GPT-3.5 baseline (dataset) & 0.56 & 0.30 \\
GPT-5.2 (Simple prompt) & 0.59 & 0.28 \\
GPT-5.2 (Chain-of-thought) & 0.60 & 0.34 \\
GPT-5.2 (CoT + Few-shot) & 0.64 & 0.32 \\
\hline
\end{tabular}
\end{table}

\begin{table}[t]
\centering
\caption{GPT-5.2 clarity-level performance using CoT + Few-Shot}
\label{tab:gpt52_clarity}
\begin{tabular}{lccc}
\toprule
\textbf{Clarity Label} & \textbf{Precision} & \textbf{Recall} & \textbf{F1-score} \\
\midrule
Clear Reply        & 0.82 & 0.42 & 0.55 \\
Ambivalent Reply  & 0.61 & 0.80 & 0.70 \\
Clear Non-Reply   & 0.51 & 0.69 & 0.58 \\
\midrule
\textbf{Overall Accuracy} & \multicolumn{3}{c}{\textbf{0.64}} \\
\bottomrule
\end{tabular}
\end{table}

\begin{table}[t]
\centering
\caption{GPT-5.2 evasion-level performance using CoT + Few-Shot}
\label{tab:gpt52_evasion}
\begin{tabular}{lccc}
\toprule
\textbf{Evasion Label} & \textbf{Precision} & \textbf{Recall} & \textbf{F1-score} \\
\midrule
Explicit              & 0.82 & 0.42 & 0.55 \\
Implicit              & 0.27 & 0.05 & 0.09 \\
Dodging               & 0.15 & 0.27 & 0.19 \\
General               & 0.21 & 0.14 & 0.17 \\
Deflection            & 0.27 & 0.16 & 0.20 \\
Partial / Half-Answer & 0.02 & 0.35 & 0.04 \\
Declining to Answer   & 0.41 & 0.63 & 0.50 \\
Claims Ignorance      & 0.69 & 0.46 & 0.55 \\
Clarification         & 0.44 & 0.91 & 0.59 \\
\midrule
\textbf{Overall Accuracy} & \multicolumn{3}{c}{\textbf{0.32}} \\
\bottomrule
\end{tabular}
\end{table}

\subsection{Overall Performance Across Tasks}

Table~\ref{tab:overall_accuracy} reports overall accuracy for clarity and evasion prediction under different prompting strategies, along with the GPT-3.5 baseline included with the CLARITY dataset.

Across all configurations, GPT-5.2 outperforms the GPT-3.5 baseline on clarity prediction. The highest clarity accuracy is achieved using chain-of-thought prompting with few-shot examples, while the highest evasion accuracy is observed with chain-of-thought prompting without few-shot examples.

\subsection{Clarity-Level Results}

Table~\ref{tab:gpt52_clarity} presents precision, recall, and F1-score for each clarity label using GPT-5.2 with chain-of-thought and few-shot prompting. The model achieves its strongest performance on Ambivalent Reply instances, indicating improved sensitivity to partial answers. Performance on Clear Non-Reply cases also improves relative to the baseline, suggesting better discrimination between partial and non-responsive answers.

\subsection{Evasion-Level Results}

Table~\ref{tab:gpt52_evasion} reports class wise evasion performance across all nine evasion categories. While chain-of-thought prompting improves overall evasion accuracy, performance varies substantially across categories. The model performs well on explicit refusal and clarification behaviors but struggles with subtle distinctions such as implicit evasion and partial answers. These results indicate that fine grained evasion classification remains challenging despite structured prompting.

\subsection{Topic Identification Results}

Topic identification accuracy under simple and chain-of-thought prompting is shown in Table~\ref{tab:topic_accuracy}. Chain-of-thought prompting improves topic accuracy from 60\% to 74\%, indicating that explicit reasoning helps the model better identify the primary issue addressed by a question–answer pair.

Despite these gains, topic identification accuracy remains below human agreement levels, particularly for semantically overlapping political domains such as economy and healthcare.

\begin{table}[h]
\centering
\caption{Topic identification accuracy across human and model based prompting strategies.}
\label{tab:topic_accuracy}
\begin{tabular}{l c}
\hline
\textbf{Model / Prompting Strategy} & \textbf{Accuracy (\%)} \\
\hline
ChatGPT – Simple Prompt & 60 \\
\hline
ChatGPT – Chain-of-Thought (CoT) & 74 \\
\hline
\end{tabular}

\end{table}

\subsection{Cross-Task Analysis}

Taken together, these results suggest a consistent pattern across tasks. Improvements in topic identification and clarity labeling indicate that reasoning oriented prompts help the model better understand question intent and response relevance. However, these gains do not consistently extend to fine grained evasion labeling, which requires more nuanced pragmatic distinctions. This divergence highlights the limits of prompt-based reasoning for subjective discourse analysis tasks.

\section{Discussion and Analysis}

This section interprets the experimental results and discusses what they reveal about prompt design for automatic clarity and evasion prediction. We focus on how different prompting strategies affect performance across label types, what limitations emerge for fine grained evasion classification, and what implications these findings have for future automatic evaluation systems.

\subsection{Effect of Prompting Strategies on Clarity Prediction}

Across all experiments, GPT-5.2 consistently outperforms the GPT-3.5 baseline on clarity prediction, indicating that newer models better align with human judgments for determining whether an answer addresses a question. More importantly, clarity accuracy increases steadily as prompts become more structured. Simple prompting yields moderate gains over the baseline, while chain-of-thought prompting produces further improvements, and the highest clarity accuracy is achieved when chain-of-thought reasoning is combined with few-shot examples.

This trend suggests that clarity prediction benefits from explicit guidance that helps the model reason about the relationship between the question and the answer. In particular, structured prompts appear to help the model better distinguish Ambivalent and Clear Non-Reply cases, which require assessing whether an answer partially addresses the question or avoids it entirely. These findings indicate that clarity is a relatively stable, high level property that can be improved through reasoning oriented prompt design. Our findings are consistent with prior work showing that model performance can be highly sensitive to prompt formulation, even when the underlying task and model remain unchanged \cite{reynolds2021prompt}, \cite{liu2023prompt}.

\subsection{Prompt Sensitivity in Evasion Prediction}

In contrast to clarity prediction, evasion prediction exhibits greater sensitivity to prompt design and does not show monotonic improvement with increased prompt complexity. While chain-of-thought prompting leads to the highest evasion accuracy among the tested configurations, adding few-shot examples does not consistently improve performance and, in some cases, slightly reduces accuracy.

This behavior reflects the inherent difficulty of evasion labeling. Evasion categories represent fine grained distinctions between different avoidance strategies, such as deflection, generalization, or partial answers. These distinctions are often subtle and context dependent, making them harder for models to identify reliably. The results suggest that while explicit reasoning steps help the model separate broad evasion types, additional examples may introduce competing patterns that do not generalize uniformly across categories.

Overall, these findings highlight that evasion prediction remains a challenging task and is more fragile than clarity prediction, even for stronger models and more sophisticated prompts.

\subsection{Error Analysis}

To better understand model behavior beyond aggregate metrics, we qualitatively analyzed cases where the predicted labels did not match human annotations. Most clarity-level errors occur at the boundary between Ambivalent Reply and Clear Non-Reply, indicating difficulty in determining whether partial information constitutes a response to the core question.

At the evasion level, the majority of errors arise between closely related categories such as Implicit, General, and Dodging. In many instances, the model correctly identifies that a response is evasive but struggles to assign the precise evasion subtype. This suggests that fine grained evasion distinctions require nuanced pragmatic and discourse level cues that are not always captured through prompt-based reasoning alone.

These observations help explain why improvements in clarity prediction do not consistently translate to gains in evasion accuracy or hierarchical exact match performance.

\section{Conclusion}
The contributions of this work lies in its controlled, prompt-centric evaluation of automatic clarity and evasion prediction. By holding the dataset, model family, and evaluation pipeline constant and varying only the prompt structure, we isolate the effect of prompting strategies on hierarchical labeling performance. Our results provide empirical evidence that increased prompt complexity does not uniformly improve performance across all labels and that reasoning based prompts benefit clarity more consistently than evasion.

These findings complement prior work on the CLARITY dataset by offering a deeper understanding of how prompt design interacts with hierarchical annotation schemes, and they highlight the need for careful, task specific evaluation of prompting techniques in automatic assessment systems.

\subsection{Implications for Automatic Evaluation Systems}

Taken together, the results demonstrate that prompt design plays a crucial role in automatic clarity evaluation, but its effects are label-dependent. reasoning based prompts reliably improve high level clarity judgments, while fine grained evasion classification remains difficult and sensitive to prompt formulation.

These findings have practical implications for the design of automatic evaluation systems. For applications that primarily require distinguishing clear from unclear responses, structured prompting can yield meaningful gains without modifying the underlying model. However, for tasks that demand reliable identification of specific evasion strategies, prompt-based approaches alone may be insufficient, and additional supervision or model level adaptations may be necessary.

\subsection{Limitations}

This study focuses on prompt-based evaluation using a single model family and a single dataset. Results may not generalize to other domains or annotation schemes. Additionally, topic identification was evaluated using a single topic label per instance, which may not capture multi topic interactions/topics in complex political discourse. These results reflect broader challenges in evaluating large language models, where accuracy metrics may not fully capture nuanced behaviors such as ambiguity, evasion, or reasoning consistency \cite{liang2022holistic}.

\end{document}